\documentclass{easychair}

\usepackage{multirow}
\usepackage{graphicx}
\usepackage{xspace}
\usepackage{booktabs} 
\usepackage{amsmath, amsfonts, amssymb, amsthm}
\usepackage[inline]{enumitem}
\usepackage{cleveref}
\crefname{subsection}{Sect.}{Sects.}
\Crefname{subsection}{Sect.}{Sects.}
\crefname{subsubsection}{Sect.}{Sects.}
\Crefname{subsubsection}{Sect.}{Sects.}

\newcommand{\ourtitle}{SAGE-5GC: Security-Aware Guidelines for Evaluating Anomaly Detection in the 5G Core Network} 
\newcommand{\featext}{\mathcal{F}}
\newcommand{\detect}{\mathcal{D}}
\newcommand{\R}{\mathbb{R}}

\newcommand{\x}{\mathbf{x}}
\newcommand{\gade}{\texttt{GA$_{\mathrm{DE}}$}\xspace}
\newcommand{\gaes}{\texttt{GA$_{\mathrm{ES}}$}\xspace}
\newcommand{\p}{\mathbf{p}}

\newtheorem{defn}{Definition}

\title{\ourtitle}

\author{
Cristian Manca\inst{1}\thanks{These authors contributed equally.}
\and
    Christian Scano\inst{1}\inst{,2}\footnotemark[1]
\and
    Giorgio Piras\inst{1}
\and
    Fabio Brau\inst{1}
\and
    Maura Pintor\inst{1}
\and
    Battista Biggio\inst{1}
}

\institute{
  University of Cagliari,
  Cagliari, Italy \and
  Sapienza University,
  Rome, Italy\\
  \email{
  name.surname@unica.it}
 }

\authorrunning{Manca et al.}

\titlerunning{SAGE-5GC}

\newcommand{\myparagraph}[1]{\smallskip \noindent \textbf{#1}}
\newcommand{\mysubparagraph}[1]{\noindent \textit{#1}}

\makeatletter
\DeclareRobustCommand\onedot{\futurelet\@let@token\@onedot}
\def\@onedot{\ifx\@let@token.\else.\null\fi\xspace}

\makeatother

\begin{document}
\maketitle

\begin{abstract}
Machine learning-based anomaly detection systems are increasingly being adopted in 5G Core networks to monitor complex, high-volume traffic. 
However, most existing approaches are evaluated under strong assumptions that rarely hold in operational environments, notably the availability of independent and identically distributed (IID) data and the absence of adaptive attackers.
In this work, we study the problem of detecting 5G attacks \textit{in the wild}, focusing on realistic deployment settings.
We propose a set of Security-Aware Guidelines for Evaluating anomaly detectors in 5G Core Network (SAGE-5GC), driven by domain knowledge and consideration of potential adversarial threats.
Using a realistic 5G Core dataset, we first train several anomaly detectors and assess their baseline performance against standard 5GC control-plane cyberattacks targeting PFCP-based network services.
We then extend the evaluation to adversarial settings, where an attacker tries to manipulate the observable features of the network traffic to evade detection, under the constraint that the intended functionality of the malicious traffic is preserved.
Starting from a selected set of controllable features, we analyze model sensitivity and adversarial robustness through randomized perturbations. 
Finally, we introduce a practical optimization strategy based on genetic algorithms that operates exclusively on attacker-controllable features and does not require prior knowledge of the underlying detection model. 
Our experimental results show that adversarially crafted attacks can substantially degrade detection performance, underscoring the need for robust, security-aware evaluation methodologies for anomaly detection in 5G networks deployed in the wild.
\end{abstract}

\section{Introduction}
\label{sec:introduction}
The 5G Core network is the central component of the 5G architecture, responsible for control and user plane functions such as authentication, mobility management, and session control. 
Its cloud-native, service-based design generates large volumes of dynamic and heterogeneous traffic that is difficult to monitor using traditional rule-based security mechanisms.
These characteristics enable flexibility and scalability, but also introduce new attack surfaces and operational challenges for network monitoring and protection. 
To protect this central component, machine learning (ML) and, in particular, anomaly detection systems are increasingly employed to analyze large volumes of heterogeneous traffic and to identify malicious activities that may not be captured by traditional rule-based or signature-based mechanisms.
Ensuring the robustness of these learning-based detectors is essential for secure network operation.
Despite their promise, most existing machine-learning approaches for anomaly detection in 5G networks are evaluated under assumptions that are difficult to satisfy in real deployments. 
In particular, they often rely on the availability of independent and identically distributed (IID) data and consider static threat models in which attackers do not adapt to the presence of learning-based defenses. 
In operational environments, however, 5G traffic exhibits strong temporal correlations, evolving service patterns, and configuration-driven changes that violate the IID assumption. 
At the same time, the integration of ML into critical network components introduces a new class of vulnerabilities. 
ML models are known to be susceptible to adversarial manipulation~\cite{biggio2018wild}, where an attacker subtly alters inputs to mislead the model with the intent of evading detection.
\begin{figure}
    \centering
    \includegraphics[width=\linewidth]{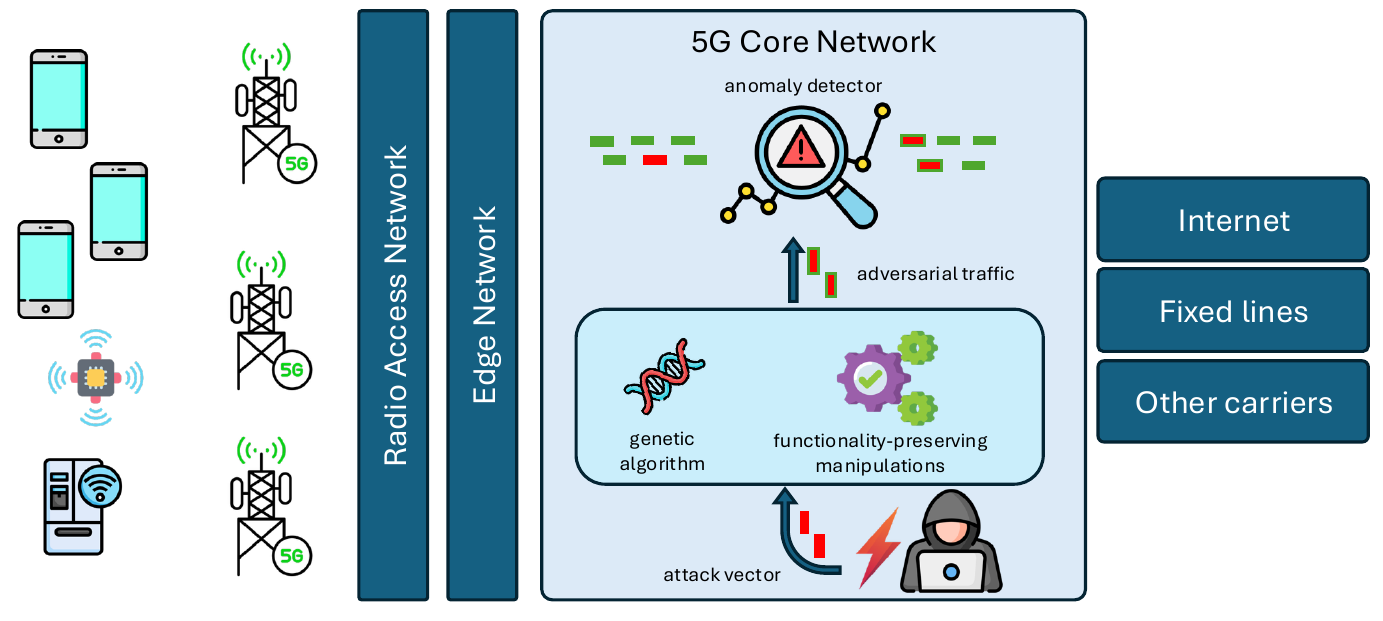}
    \caption{An overview of our proposed realistic SAGE-5GC evaluation, which confronts anomaly detection models with attackers that manipulate traffic to evade the system.}
    \label{fig:overview}
\end{figure}
This gap between evaluation practices and real-world conditions raises concerns about the reliability of current anomaly detection systems when deployed \textit{in the wild}. 
High detection accuracy reported under controlled experimental settings does not necessarily translate into robust performance in the presence of traffic drift, non-stationarity, and adaptive attackers. 
In particular, adversarial attacks against machine-learning–based detectors pose a significant threat to system integrity, as they aim to induce misclassification through subtle, protocol-compliant feature manipulations.
In this work, we address these challenges by studying anomaly detection in realistic 5G Core network scenarios. 
We propose \textbf{SAGE-5GC}, a set of \emph{Security-Aware Guidelines for Evaluating anomaly detection in the 5G Core network}, driven by domain knowledge of 5G protocols and operational constraints, and explicitly accounting for adversarial threats. 
Using a realistic 5G Core dataset, we first evaluate the baseline performance of several anomaly detection models against \textit{traditional} cyberattacks targeting control-plane network services (PFCP). 
We then extend the analysis to adversarial settings, in which attackers preserve the semantics and effectiveness of malicious traffic while manipulating observable features to evade detection. 
We study the sensitivity of the models to feature perturbations and assess their robustness under both random and optimized adversarial strategies. 
To achieve the latter, we introduce a practical, model-agnostic optimization approach based on genetic algorithms that operates solely on attacker-controllable features and does not require prior knowledge of the underlying detection model. 
An overview of our method is presented in \Cref{fig:overview}, which depicts the attacker-in-the-loop setting considered in a realistic 5G Core environment. Starting from a malicious attack vector targeting 5G Core control-plane traffic, the adversary applies functionality-preserving and protocol-compliant manipulations to a restricted set of attacker-controllable features through a genetic–algorithm–based strategy. The resulting adversarial samples are then fed to the anomaly detector to assess its robustness.
The results of our experimental evaluation demonstrate that adversarially crafted attacks can significantly degrade detection performance, even for models that perform well under conventional evaluation settings. 
These findings highlight the need for adopting security-aware and adversarially informed evaluation methodologies for anomaly detection in 5G Core networks, particularly when systems are deployed in real-world operational environments. Our contributions are the following:

    \begin{enumerate}
        \item We provide a set of security-aware guidelines for training and testing anomaly detection systems for the 5G Core network, explicitly considering adversarial threats;
        \item We evaluate a diverse set of anomaly detection algorithms on a realistic 5G Core dataset representative of real deployment traffic; and
        \item We extend the evaluation to adversarial scenarios, in which the attacker actively manipulates features to evade detection while keeping the malicious activity functional and effective.
    \end{enumerate}

\noindent In addition, the framework code is available at \url{https://github.com/pralab/sage-5gc}.

\section{Anomaly Detection in 5G Core Networks}
\label{sec:intrusion_detection}
In this section, we first provide an overview of the 5GC network and its vulnerabilities and formalize the task of anomaly detection (\Cref{sect:anon_dect_overview}). We then describe the anomaly detection algorithms relevant to our work (\Cref{sect:anon_dect_algos}), and conclude with a set of guidelines to guide the training of these systems for deployment in a realistic and applicable scenario (\Cref{sec:training_guidelines}).

\subsection{Overview}\label{sect:anon_dect_overview}

\myparagraph{Overview of the 5G Core network.} 
The 5G Core (5GC) network is the central component of the 5G architecture, responsible for key control-plane and user-plane functions, including authentication, mobility management, session establishment, and service orchestration. 
The 5GC adopts a service-based, cloud-native architecture built on Network Function Virtualization (NFV) and edge-cloud components. 
While this design enables flexibility and scalability, it also generates high-volume, heterogeneous, and highly dynamic traffic, making continuous and automated monitoring essential for maintaining network security.

\myparagraph{Attacks against the 5GC network.} The 5GC is exposed to a wide range of traditional cybersecurity attacks targeting network services and protocols, including control-plane signaling protocols such as PFCP. 
These include denial-of-service (DoS) attacks, network scanning, and reverse shells (for lateral movement), as well as attacks that exploit vulnerabilities specific to the 5GC. 
Such attacks aim to disrupt service availability, degrade network performance, or compromise user data. 
We defer the description of these PFCP-based attacks to \Cref{sec:experiments}, where we provide details of the dataset used.

\myparagraph{Anomaly detection.} Anomaly detection provides a natural approach for monitoring the 5GC in scenarios where comprehensive labeling of malicious traffic is not available. 
The general pipeline for anomaly detection in the 5GC consists of:  
(i) raw traffic collection from core network functions (i.e., collecting packets at the transport level),  
(ii) feature extraction and preprocessing,  
(iii) training an anomaly detection model on benign traffic, and  
(iv) evaluating new incoming traffic to identify deviations from normal behavior.

In this work, we consider anomaly detection in the one-class learning setting, where the training data consists primarily of benign 5GC traffic, i.e., the detector learns a representation of normal behavior and identifies deviations as potential attacks. 
The anomaly detection is performed on features extracted from packets obtained by raw network traffic captures. 


Packets are represented as byte strings of variable length over an alphabet $[256]$, where $[n] := {1, \ldots, n}$ for $n \in \mathbb{N}$. 
A feature extraction function $\featext : [256]^* \rightarrow \mathbb{R}^d$ maps each packet $\p$ into a finite-dimensional feature space suitable for learning-based analysis. 
The resulting feature vector $\x = \featext(\p)$ is non-homogeneous and can be decomposed as $\x^\top = [\x_1^\top, \x_2^\top] \in \mathbb{N}^{d_1} \times \mathbb{R}^{d_2} \subseteq \mathbb{R}^d$, where $\x_1$ and $\x_2$ represent categorical and numerical features, respectively.
An anomaly detector $\detect : \mathbb{R}^d \rightarrow \mathbb{R}$ assigns an anomaly score to each feature vector, with higher values indicating greater deviation from learned normal behavior. An observation is classified as anomalous if its score exceeds a threshold $\tau$, i.e., if $\detect(\featext(\p))>0$.

\subsection{Anomaly Detection Algorithms}\label{sect:anon_dect_algos}
Algorithms for anomaly detection can be categorized into different families based on their underlying principles and detection mechanisms. 
We evaluate representative methods from each category to assess their effectiveness in the selected scenario.

\myparagraph{Statistical methods (\texttt{HBOS}, \texttt{COPOD}, \texttt{ECOD}) \texttt{Feature Bagging}.} 
    These approaches assume that normal data can be effectively characterized by probabilistic or statistical models, and anomalies are identified as rare or unlikely observations under such models. 
    Common strategies include constructing feature-wise histograms or fitting empirical distributions. 
    Examples within this family are the Histogram-Based Outlier Score~(\texttt{HBOS})~\cite{goldstein2012histogram}, which scores anomalies based on individual feature histograms, Copula-Based Outlier Detection~(\texttt{COPOD})~\cite{li2020copod}, which captures feature dependencies through copula models, and Empirical-Cumulative-distribution-based Outlier Detection~(\texttt{ECOD})~\cite{li2022ecod}, which relies on empirical cumulative distribution functions to estimate tail probabilities without strict parametric assumptions.
    Finally, \texttt{Feature Bagging}~\cite{yang2020feature}, combines several base detectors trained on randomly selected feature subsets, capturing attack patterns that might be missed by any single view of the data.
    
\myparagraph{Density-based techniques (\texttt{kNN}, \texttt{LOF}, \texttt{IForest}, \texttt{LODA}, \texttt{INNE}).} 
    In this category, the underlying intuition is that normal data points reside in dense regions of the feature space, while outliers or attacks are typically situated in sparser areas.
    Models adopt various ways to quantify local density or sample isolation. 
    Typical approaches in this family are $k$-Nearest Neighbors~(\texttt{kNN})~\cite{angiulli2002fast}, which evaluates the distance to neighboring points, and Local Outlier Factor~(\texttt{LOF})~\cite{breunig2000lof}, which compares the local density of a sample to those of its neighbors to identify context-dependent anomalies. 
    More scalable or ensemble-based variants, such as Isolation Forest~(\texttt{IForest})~\cite{liu2008isolation}, which isolates samples using random splits, Lightweight On-line Detector of Anomalies~(\texttt{LODA})~\cite{pevny2016loda}, which projects data onto low-dimensional subspaces, and Isolation-based anomaly detection using Nearest-Neighbor Ensembles~(\texttt{INNE})~\cite{bandaragoda2018isolation}, which combines isolation principles and nearest neighbor distances, further enhance robustness in high-dimensional scenarios.
    
\myparagraph{Geometric-based techniques (\texttt{PCA}, \texttt{ABOD}, \texttt{GMM}).} 
    These methods focus on capturing the overall shape, spread, or geometric boundaries of the normal data distribution. 
    They flag anomalies as points that fall outside of this learned structure. 
    Among the commonly used strategies are Principal Component Analysis~(\texttt{PCA})~\cite{shyu2003novel}, which identifies deviations from the principal axes of variation via reconstruction errors; Angle-Based Outlier Detection~(\texttt{ABOD})~\cite{kriegel2008angle}, which analyzes angular variance between points, particularly suited for high-dimensional data, and Gaussian Mixture Models~(\texttt{GMM})~\cite{aggarwal2016probabilistic}, which probabilistically cluster data and detect outliers as those with low likelihood under the learned mixture components.
    
\myparagraph{Ensemble techniques.} 
    This family comprises methods that aggregate the outputs of multiple anomaly detectors, often trained on different feature subsets or with different initializations, to improve reliability and generalization.

\subsection{Guidelines to Train Real-time Anomaly Detectors}
\label{sec:training_guidelines}

To be effectively deployed, an anomaly detection system for the 5GC must satisfy the following guidelines for training (GT), which we summarize here and detail below:
\begin{itemize}
    \item[GT1] \textit{Environmental independence}, the system should rely exclusively on protocol and behavioral features, explicitly removing environment-dependent information such as IP addresses, port numbers, and deployment-specific identifiers.
    \item[GT2] \textit{Protocol-specific design}, feature selection should be guided by domain knowledge of the targeted 5GC protocols and attack surfaces, eliminating irrelevant traffic and redundant fields.
    \item[GT3] \textit{Robust feature representation}, all retained features must be systematically transformed into stable, numeric representations suitable for machine learning, with careful handling of missing values and heterogeneous scales.
    \item[GT4] \textit{Anomaly-agnostic training}, malicious packets should therefore be excluded from the training phase and used only for validation and testing.
\end{itemize}

\myparagraph{GT1. Environmental independence.} Data representation in the network traffic domain plays a critical role in the performance of anomaly detection methods in machine learning~\cite{yang2020feature}. 
Therefore, it is essential to avoid any representation that relies on specific packet header values, such as the value of a source or destination IP address or port number, the value of a protocol (e.g., TCP, UDP), and so on, to avoid developing models that would define normal behavior as specific to a particular network connection point (e.g., IP address or port). 
This is important because training models based on specific values, such as IP addresses or port numbers, would simply define anomalies as any deviation in network destinations from those detected in the training data. 
Consequently, a model trained on traffic traces collected in one part of the network would be unable to perform general novelty detection when applied or deployed to other parts of the network. 
Furthermore, the model would fail to detect new behaviors originating from the same IP addresses over time, as would happen if a device were compromised or changed its behavior as a result of changes in the environment or in the way users interact with the device. 
For similar reasons, it is important to avoid relying on port numbers. 
While a significant amount of traffic, both normal and abnormal, traverses common ports (for example, port 443 or HTTPS), client traffic, in contrast, originates from ports that are constantly changing (for example, increasing). 
Defining normal traffic based on client ports would result in a significant number of false positives, simply due to the normal behavior of clients originating connections from different ports over time.

\myparagraph{GT2. Protocol-specific design.} 
Since the focus is on control-plane security in the 5GC, attention should be restricted to the Packet Forwarding Control Protocol (PFCP), which governs session and forwarding state between core network functions and is exclusively transported over UDP. 
Traffic unrelated to this control-plane interaction, such as TCP- and ICMP-based packets, should therefore be excluded at the packet level, and all features associated with these protocols should be removed. 
This restriction ensures that the analysis focuses on the protocol layers that are directly relevant to control-plane attack scenarios, namely IP, UDP, and PFCP, while avoiding the introduction of noise from unrelated transport or application-layer behaviors.

\myparagraph{GT3. Robust feature representation.}
To ensure that the feature space remains both minimal and informative, constant, empty, and redundant protocol fields should be removed. 
This includes features that take a single unique value across all observations or encode no meaningful variability. Such fields do not contribute to anomaly detection and may introduce unnecessary noise or bias. Features that present missing values, instead, should be explicitly handled and imputed when necessary, since many anomaly detection models cannot operate on feature vectors containing null values, while non-numerical features should be transformed into a suitable numerical representation.

Finally, feature scaling should be considered as a recommended preprocessing step, particularly in the presence of heterogeneous feature ranges and extreme values, which are common in network traffic data. 
Robust scaling methods that rely on distribution quantiles can mitigate the influence of outliers without assuming normality. 
Although scaling is not strictly required for all anomaly detection models, its impact should be explicitly evaluated to ensure a transparent and reproducible assessment.

\myparagraph{GT4. Anomaly-agnostic training.} Malicious packets should be excluded from the training phase of anomaly detection models and used exclusively for validation and testing. Including malicious traffic during training undermines the novelty-detection objective and leads to overly optimistic performance estimates that do not reflect real deployment conditions, where attacks are unknown at training time.

\section{Attacking Anomaly Detectors for the 5GC  Network}
\label{sec:PentAttack}


Traditional cyberattacks on 5GC networks aim to compromise regular system functionality, regardless of the presence of ML-based defenses.
In contrast, adversarial attacks explicitly target the integrity of ML models.
However, directly applying gradient-based adversarial methods such as the Projected Gradient Descent (PGD)~\cite{madry2018towards} to packet-level feature representations is generally infeasible~\cite{catillo2024towards}.
PGD perturbs input features in the direction of the loss gradient and enforces an $\ell_p$ constraint to maintain the perturbation small.
While this approach works well with images, applying PGD to the packet features without accounting for protocol semantics is not suitable for our case.
In fact, when applied to features derived from PFCP control-plane traffic, such perturbations may result in invalid protocol fields, inconsistent flag combinations, or malformed message structures that violate 5G specifications. 
As a consequence, the modified packets would no longer be accepted by network functions, thereby breaking the attack's functionality and thus invalidating the adversarial objective.
Therefore, our goal is to test the detectors when they face malicious traffic that is subtly modified to appear benign to the detector while still performing the intended harmful activity.
In this section, we first outline the formalization (\Cref{sec:adv}) and detail the solving algorithms (\Cref{sec:attack_optim}) of our attack. Then, we draw the guidelines for evaluating anomaly detection for the 5GC network (\Cref{sec:testing_guidelines}).

\subsection{Adversarial Attacks against Anomaly Detectors for 5GC Network}
\label{sec:adv}

To formalize the attack, we first identify our assumptions and then provide definitions that enable the attack to occur at the packet level.
We build on these observations to design our strategies for finding adversarial perturbations that are both feasible and compliant with the protocol, first relying on a random search, and then on a more efficient genetic algorithm.

\myparagraph{Threat Model.} 
We assume a model-agnostic attacker, capable of injecting or spoofing PFCP traffic in the 5GC network. 
From an evaluation perspective, assuming a model-agnostic attacker provides a conservative and robust assessment of anomaly detection systems. 
Model-specific adversaries may exploit vulnerabilities tied to a particular model, leading to conclusions that do not generalize across different detection models and complicate comparisons. 
In contrast, a model-agnostic attacker relies solely on observable behavior and feature-level knowledge, reflecting realistic deployment scenarios in which the internal details of the detector are not exposed.

We make the following assumptions. The attacker: 
\begin{enumerate*}[label=(\roman*)]
    \item has no access to the internal parameters or architecture of the anomaly detector $\detect$;
    \item can observe the detector's outcome, such as identifying which traffic is flagged as anomalous, and the detection score (therefore, has also access to the threshold $\tau$ that can be, in practice, inferred from a set of triggered detections and the observed scores); and
    \item has access to the feature extractor $\featext$.
\end{enumerate*} 

The third assumption enables the adversary to operate in the feature space while deriving the constraints required to map admissible perturbations back to the input (packet) space, ensuring that the resulting attacks are feasible in practice.
Feasible modifications are restricted to traffic fields whose alteration does not violate protocol compliance or interfere with the core functionality of the malicious activity. 
This reflects realistic constraints in attacks on the 5GC network, where attack effectiveness must be preserved while remaining compliant with protocol specifications. 
A formalization of these constraints is provided in the remainder of this section.

\begin{defn}[Feasibility]
Let $J \subseteq [d]$ be a set of feature indices.  
We say that a modification constrained to the features in $J$ is \emph{feasible} if it produces a realizable feature vector. Formally, for each packet $\p$ with feature representation $\x = \featext(\p) \in \mathbb{R}^d$, any modified feature vector $\x' \in \mathbb{R}^d$ such that $\x'_i = \x_i \quad \text{for all } i \notin J$ is said to be feasible if the modification is \emph{invertible}, i.e., if there exists a packet $\p'$ such that $\featext(\p') = \x'$. With a slight abuse of notation, we say that $J$ is feasible if it allows for feasible modifications.
\end{defn}

\begin{defn}[Compliance]
    We say that two packets $\p, \mathbf{q}$ are compliant (namely $\p\equiv \mathbf{q}$) if $\p$ preserves protocol compliance and attack integrity (or functionality) of $\mathbf{q}$.
\end{defn}

Let $J\subseteq[d]$ be a set of feasible features, given a malicious packet $\p$, which we assume is successfully detected, i.e.,  $\detect(\p)\ge\tau$, the aim of the attacker is to deduce a new packet $\p'\equiv\p$ which preserve the functionality of $\p$ but such that $\detect(\p')<\tau$. Formally, this involves solving the following minimization problem (MP):
\begin{equation}
    \min_{\p'\equiv \p} \left[\detect(\featext(\p'))-\tau\right]_+,\quad \text{s.t.}\quad \featext(\p)_i=\featext(\p')_i,\,\forall i\not\in J,
    \label{eq:minP}
    \tag{MP}
\end{equation}
where $[\cdot]_+$ indicates the positive part of a function, $\tau$ is the detection threshold.

\subsection{Attack Optimization Algorithms}\label{sec:attack_optim}

We emphasize that we specifically avoid minimization methods that leverage the gradient of $\detect$ to maintain the model-agnostic nature of the attack. Hence, we tackle Problem~\ref{eq:minP} through two algorithms, namely \textit{random search} and \textit{genetic algorithm}.
Finally, we note that throughout the paper, we assume the set of feasible indices $J$ is manually selected by the attacker, based on their previous knowledge of the network system, and is therefore not optimized. \medskip

\myparagraph{Random search (\texttt{RS}).} 
In the random attack strategy, each malicious sample is perturbed by randomly modifying only the set of feasible features that produce a compliant packet, thus, those realizable modifications permitted by protocol constraints and that preserve the integrity of the attack. 
During a preparatory stage, a feasible set $J$ is selected, and, for each feature $j\in J$, the distributions $\mu_j$ are deduced on previously spoofed data. 
The minimum problem is then estimated by considering $\p' = \featext^{-1}(\x')$ where $\x'_j\sim\mu_j$ for each $j\in J$ and $\x'_j=\x_j=\featext{(\p)}_j$ otherwise, and sampling $\p'$ until the objective of \ref{eq:minP} reaches $0$.


\myparagraph{Genetic algorithms (\texttt{GA}).} We also address Problem~\ref{eq:minP} by means of a genetic algorithm (GA), which is well suited for model-agnostic optimization under feasibility constraints, such as feasibility and compliance. 
The algorithm iterates by generating individuals in the population, which represent a candidate packet $\p'$ through its feature vector $\x'=\featext(\p')\in\R^d$, where genes corresponding to indices in $J$ are free to vary, while all remaining genes are fixed, i.e., $x'_i=x_i$ for all $i\notin J$. 
By construction, individuals are restricted to feasible modifications, and the decoding step $\p'=\featext^{-1}(\x')$ ensures that, if inverted, these manipulations can correspond to real packets.
The algorithm iterates the following steps: 
\begin{enumerate}[label=(\roman*)]
\item The initial population is generated by sampling features in $J$ according to the empirical distributions $\mu_j$ introduced above, while keeping all other features unchanged.
\item  Given a population $\{\p'_k\}_{k=1}^N$, the fitness of each individual is defined as
\[
f(\p'_k)=\left[\detect(\featext(\p'_k))-\tau\right]_+,
\]
which directly mirrors the objective in~\eqref{eq:minP}. Lower fitness values correspond to more successful evasion attempts.
\item At each generation, individuals are selected according to their fitness and combined through crossover operators acting only on the indices in $J$, so as to preserve feasibility and compliance. 
Mutation is then applied by randomly resampling a subset of features in $J$ from the corresponding distributions $\mu_j$, while leaving all indices $i\notin J$ untouched.
\item The algorithm iterates until either an individual with zero fitness is found, corresponding to a successful evasion, or a maximum number of generations is reached. 
In this way, the GA explores the space of feasible and compliant modifications more efficiently than pure random sampling, while remaining fully compatible with the model-agnostic threat model assumed throughout this work.
\end{enumerate}

\subsection{Guidelines for Evaluating Anomaly Detectors}
\label{sec:testing_guidelines}
Evaluating anomaly detection systems in the 5G Core network requires consideration of several implementation constraints, including novelty-based operation, class imbalance, and adaptive attacks. 
To this end, we define a set of guidelines for evaluating anomaly detectors (EGs), which complement the training guidelines provided in \Cref{sec:training_guidelines} and outline a principled framework for evaluating both the effectiveness and robustness of detection in realistic 5G scenarios. 
Again, we summarize these guidelines here and discuss them in detail below:
\begin{itemize}
    \item[GE1] \textit{Realistic data and novelty-driven setup,} evaluation must reflect realistic 5GC network conditions, enforcing novelty-detection assumptions and preserving the natural imbalance between benign and malicious traffic.
    \item[GE2] \textit{Alignment with imbalanced and security-critical settings,}  evaluation should rely on performance indicators that capture detector behavior under class imbalance and heterogeneous attack patterns, combining global measures of detection capability with class-level analyses to reveal security-relevant weaknesses.
    \item[GE3] \textit{Adversarial robustness,} 
    since anomaly detectors in the 5GC operate in adversarial environments, evaluation must explicitly assess robustness against adaptive attackers. 
    This includes testing the detectors against adversarial manipulations to ensure their robustness.
\end{itemize}
\myparagraph{GE1. Realistic data and novelty-driven setup.}
Evaluation protocols should reflect the conditions under which anomaly detectors are expected to operate once deployed in a real 5GC network. 
In practice, this implies adopting a strict novelty-detection setting, where models are trained primarily or exclusively on benign traffic and are required to identify deviations corresponding to previously unseen attacks. 
Benign traffic used for training and evaluation should originate from distinct flows, clients, or operational instances, rather than from overlapping or trivially related samples.
For instance, testing on packets from the same PFCP sessions observed during training leads to near-perfect performance by construction, rather than genuine anomaly detection.
Moreover, evaluation datasets should preserve the strong class imbalance that characterizes real 5GC environments, where malicious events are rare compared to legitimate traffic. 
By enforcing these characteristics, evaluation results more accurately reflect the detector’s ability to generalize and operate effectively in real deployment scenarios.

\myparagraph{GE2. Alignment with imbalanced and security-critical settings.}
In security-critical environments such as the 5GC, evaluation must capture detector behavior beyond aggregate performance summaries. 
Due to the heterogeneity of attack types and their uneven prevalence, a detector may appear effective overall while systematically failing to detect specific attacks or generating excessive false alarms on benign traffic. 
A good practice in this scenario is to achieve the best trade-off between detection and false alarms.
For this reason, evaluation should adopt performance indicators that jointly characterize global detection capability and class-level behavior. 
This allows practitioners to identify security-relevant weaknesses, such as the poor detection of specific attack categories or unstable behavior across traffic conditions, which would otherwise remain hidden when relying solely on aggregate measures.

\myparagraph{GE3. Adversarial robustness.}
Anomaly detectors deployed in the 5GC must be evaluated not only against traditional cybersecurity attacks, but also under adversarial conditions in which attackers actively manipulate traffic to evade the learning-based detectors. 
Since these systems operate in a security-sensitive context, robustness to adversarial behavior is a fundamental evaluation criterion. 
Evaluation should therefore include the assessment of detector robustness against adversarial manipulations that respect protocol compliance and preserve the functionality of the attack. 
Perturbations must be limited to attacker-controllable features and reflect realistic constraints imposed by 5G Core protocols and attack semantics. 
This form of adversarial evaluation helps distinguish detectors that are robust by design from those whose performance relies on fragile correlations in the feature space.

\section{Experiments}
\label{sec:experiments}
Following the structure of the previous section, we report in~\Cref{sec:exp_setup} the experimental setup, we discuss in \Cref{sect:results_detectors} the evaluation of the detector's performance on clean data, while in \Cref{sect:results_attacks} we discuss the evaluation of the detectors under the adversarial attacks formulated in \Cref{sec:adv}.
\subsection{Experimental Settings}
\label{sec:exp_setup}
In this section, we present the experimental settings adopted during the experiment phase and discuss the results obtained.

\myparagraph{Dataset.}
In our experiments, we use 5G-Attacks~\cite{5gattacks} a recently published dataset containing several attack categories targeting the 5G Core network.\footnote{\url{https://github.com/clem272001/5G-Attacks}} 
This dataset provides labeled network traffic, including realistic attack scenarios, making it suitable for machine learning-based intrusion detection solutions.
The resulting datasets include 5G-oriented attack scenarios that exploit specific vulnerabilities in next-generation mobile networks. 
An overview of the dataset is reported below (see \Cref{app:datasets} for further details).
The dataset includes five representative PFCP-based attacks targeting the 5GC. 
All attacks exploit the absence of authentication, integrity protection, and strict input validation in PFCP signaling, and are generated by manipulating protocol fields while preserving syntactic validity.\medskip

\mysubparagraph{PFCP Restoration-TEID.}
This attack disrupts PFCP session recovery by injecting forged restoration messages with out-of-range TEID values and corrupted F-TEID parameters. 
The resulting desynchronization between control-plane and user-plane tunnel state leads to traffic misrouting, session drops, or UPF crashes.\medskip

\mysubparagraph{PFCP Flood.}
A control-plane denial-of-service attack that overwhelms PFCP endpoints with unsolicited Session Establishment and Heartbeat messages. 
Randomized identifiers and sequence numbers lead to excessive state processing, exhausting control-plane resources and degrading session management performance.\medskip

\mysubparagraph{PFCP Deletion.}
This attack forges Session Deletion Requests targeting active SEIDs, resulting in the premature removal of PDRs, FARs, and tunnel state. The immediate teardown of sessions results in the abrupt interruption of user-plane traffic.\medskip

\mysubparagraph{PFCP Modification.}
Counterfeit Session Modification messages are injected to corrupt forwarding behavior at the UPF. 
By disabling forwarding actions and altering encapsulation and interface parameters, the attack induces packet drops, misrouting, or invalid tunneling.\medskip

\mysubparagraph{UPF PDN-0 Fault.}
This attack exploits the insufficient validation of PDN Type~0 session contexts by inducing inconsistent PFCP state through malformed identifiers and invalid F-TEID flag combinations. 
The resulting inconsistencies cause session establishment failures, unstable state transitions, or traffic misrouting.\medskip

\begin{table}[t]
\centering
\renewcommand{\arraystretch}{1.3}
\begin{tabular}{l|c|c|c}
\hline
\textbf{Characteristics} & \textbf{Train Dataset} & \textbf{Validation Dataset} & \textbf{Test Dataset} \\
\hline
\multicolumn{4}{c}{\textbf{Packets}} \\
\hline
Normal                & 21341 & 4731 & 4732 \\  
PFCP Restoration-TEID & 0     & 13 & 22   \\    
PFCP Flood            & 0     & 1039 & 1026 \\  
PFCP Deletion         & 0     & 7 & 13   \\     
PFCP Modification     & 0     & 16 & 12   \\    
UPF PDN-0 Fault       & 0     & 10 & 12   \\    
\hline
\end{tabular}
\caption{Sample Distribution of Train, Validation, and Test datasets.}
\label{tab:dataset_distribution}
\end{table}

The raw datasets extracted via \texttt{tshark} contain multi-protocol packet-level information spanning IP, UDP, TCP, and the 5GC-specific PFCP protocol. 
Features include packet-level attributes (e.g., header lengths, flags, timestamps), protocol-specific fields (e.g., PFCP TEID values), and source/destination data such as IP addresses and transport-layer port numbers. 
The original dataset consists of three specific splits: one for unsupervised training, one for supervised training, and the last for testing.
The detailed sample distribution for each attack class and dataset partition is reported in \Cref{tab:dataset_distribution}.

\begin{table}[t]
    \centering
    \begin{tabular}{l|c|c}
    \hline
    \textbf{Protocol} & \textbf{Before Preprocessing} & \textbf{After Preprocessing} \\
    \hline
    \multicolumn{3}{c}{\textbf{Number of Features}} \\
    \hline
    IP   & 16 & 4   \\
    UDP  & 9 & 1 \\
    PFCP & 45 & 28 \\
    \hline
    \end{tabular}
    \caption{Number of Features before and after applying Preprocessing}
    \label{tab:features_by_protocol}
\end{table}

\textit{Following GT4, we have to modify the setup to perform anomaly detection training in an unsupervised manner, providing only the benign traffic.}
Starting from the cleaned dataset version provided with the original 5G-Attacks repository, we construct our own train and test splits by merging the entirety of dataset 1 with all legitimate samples (i.e., those without an attack label) from dataset 2, while our test set comprises all labeled attack samples from dataset 2 together with all samples from dataset 3. 

\textit{Following GT2, we need to remove irrelevant traffic from the dataset.}
To focus our analysis on relevant control-plane traffic, and in line with the focus on PFCP-based attacks, we further filter out all TCP and ICMP packets, retaining only UDP flows.
\Cref{tab:features_by_protocol} summarizes the feature distribution across protocols (excluding TCP, since PFCP communication in the 5GC relies on UDP) before and after applying our preprocessing pipeline, reducing the feature space and preserving only the relevant and discriminative information for anomaly detection.

\textit{In line with GT3, missing values were imputed using robust and well-established methods from the \texttt{scikit-learn} library~\cite{scikit-learn}}: For categorical features, we apply the \texttt{SimpleImputer} with the most frequent category strategy, ensuring consistent treatment of common protocol or flag fields; for numerical features, we use the \texttt{IterativeImputer} with a \texttt{Random Forest Regressor} as estimator to exploit correlations among features and generate plausible imputations.

These choices help maintain the statistical properties of the data and reduce the risk of introducing bias or spurious patterns in subsequent modeling steps.   
Finally, we apply the \texttt{RobustScaler} from \texttt{scikit-learn} independently to each numerical feature. The resulting dataset is a compact, protocol-based, and environment-independent representation of 5G traffic.\medskip

\myparagraph{Models.}
We consider a diverse set of state-of-the-art anomaly detectors, imported from the \texttt{PyOD} library~\cite{zhao2019pyod}, as well as custom pipelines tailored for the 5G network data. Specifically, we evaluate: 
\texttt{ABOD}~\cite{kriegel2008angle}, \texttt{COPOD}~\cite{li2020copod}, \texttt{ECOD}~\cite{li2022ecod}, \texttt{FeatBagg}~\cite{yang2020feature}, \texttt{GMM}~\cite{aggarwal2016probabilistic}, \texttt{HBOS}~\cite{goldstein2012histogram}, \texttt{Isolation Forest}~\cite{liu2008isolation}, \texttt{INNE}~\cite{bandaragoda2018isolation}, \texttt{kNN}~\cite{angiulli2002fast}, \texttt{LODA}~\cite{pevny2016loda}, \texttt{LOF}~\cite{breunig2000lof}, and \texttt{PCA}~\cite{shyu2003novel}. 
For each detector, we tune the hyperparameters via grid search on the validation set.
For each model, the grid search spans detector-specific parameters such as neighborhood size for density-based methods (e.g., k in \texttt{kNN}/\texttt{LOF}), number of bins or splits for histogram/statistical models (e.g., \texttt{HBOS}), and contamination (expected outlier proportion).

We also use ensemble models, which exploit the complementarity of heterogeneous detection strategies. 
In each ensemble, the scores deduced from the outcome of a subset of base detectors are fed into a binary classifier (specifically, a Support Vector Machine) to deduce an aggregated output score. The ensemble-model parameters are trained on the validation set, while the parameters of the base detectors are kept fixed. \textit{Following the GT4 guideline, the validation set is free of malicious packets.}
All ensembles adopt a Support Vector Classifier (\texttt{SVC}) with an RBF kernel, implemented using the \texttt{scikit-learn} python library.
The hyperparameters of the \texttt{SVC} (i.e., the regularization parameter $C$ and the scaling factor $\gamma$) are selected on different runs to maximize the F1-score on the validation set.
We consider the following  ensemble models: 
\begin{itemize}
    \item \texttt{Ens-HKAIP}: \texttt{SVC} with $C = 10$ and $\gamma = 10$, combining \texttt{HBOS}, \texttt{KNN}, \texttt{ABOD}, \texttt{INNE}, and \texttt{PCA};
    \item \texttt{Ens-HKGIP}: \texttt{SVC} with $C = 10$ and $\gamma = 10$, combining \texttt{HBOS}, \texttt{KNN}, \texttt{GMM}, \texttt{INNE}, and \texttt{PCA};
    \item \texttt{Ens-HKLIP}: \texttt{SVC} with $C = 10$ and $\gamma = 10$, combining \texttt{HBOS}, \texttt{KNN}, \texttt{LOF}, \texttt{INNE}, and \texttt{PCA};
    \item \texttt{Ens-HKLIF}: \texttt{SVC} with $C = 100$ and $\gamma = 100$, combining \texttt{HBOS}, \texttt{KNN}, \texttt{LOF}, \texttt{INNE}, and \texttt{FeatBagg}.
\end{itemize}
We expect this design to enable the system to compensate for the weaknesses of individual detectors in specific attack categories, leading to more stable and accurate anomaly detection performance compared to standalone models.

\textit{Following GT3, we evaluate also the effect of scaling the features.} Therefore, we measure results in two cases, i.e., with and without the application of feature scaling using \texttt{RobustScaler} applied to the features (fitted on the training set). Reporting both configurations clarifies the impact of the scaling on the detection performance and its influence on model stability and robustness. 

\myparagraph{Evaluation metrics.}
To assess the effectiveness of the detection, we employ a set of evaluation metrics commonly used in anomaly detection and classification tasks. 
The recall, which quantifies the proportion of attack samples correctly identified as anomalies. 
Precision measures the proportion of samples flagged as anomalous that correspond to actual attacks, reflecting the cost of false positives.
The F1-score, defined as the harmonic mean of precision and recall, summarizes the trade-off between detection effectiveness and false alarm rate.
Finally, the Area Under the Receiver Operating Characteristic curve (AUC) evaluates the model’s ability to discriminate between benign and malicious traffic across all possible detection thresholds, independently of a specific operating point.

\myparagraph{Attack Algorithms.}
For reproducibility, we report the hyperparameters used in the implementation of the attacks.
The random search \texttt{RS} attack is performed by sampling $\p'$ a single time, without attempting multiple attacks on the same packet. 
For the Genetic Algorithm (\texttt{GA}), we leverage population-based evolutionary optimization using the \texttt{Nevergrad} library.\footnote{\url{https://facebookresearch.github.io/nevergrad/}} We use two algorithms:
\medskip

\mysubparagraph{Differential Evolution (\gade).} 
The optimizer is instantiated with a population size of $20$ (\texttt{popsize=20}), a two-point crossover scheme (\texttt{crossover="twopoints"}), and ensures heritage propagation (\texttt{propagate\_heritage=True}). 
For every attack sample, the search proceeds for a fixed query budget of 100, thus allowing up to 100 candidate adversarial samples can be generated and tested against the detector.\medskip

\mysubparagraph{Evolution Strategy (\gaes).} 
The alternative optimizer, available by switching the relevant code block, also uses a population size of $20$ (\texttt{popsize=20}) and a recombination ratio of $0.9$, with a query budget of $100$. 
All other parameters are set as defaults.\medskip

Both algorithms evolve a population of adversarial samples, but their internal logic for generating new candidates differs. \gade relies on combining individuals via differential perturbations and explicit crossover; thus, new solutions are created by adding scaled differences between members of the current population, with two-point crossover further mixing feature values. \gaes, in contrast, emphasizes random mutation and optional recombination of ``parent'' solutions, focusing on stochastic local sampling of the search space. 
For all runs, we set the random seed to 42 to ensure experiment reproducibility. Each optimization process is further bounded by a strict query budget per attack sample, guaranteeing that all experiments are both controlled and repeatable across all attack types.

\subsection{Evaluation of Detectors}\label{sect:results_detectors}
\begin{figure}[t]
    \centering
    \includegraphics[width=1\linewidth]{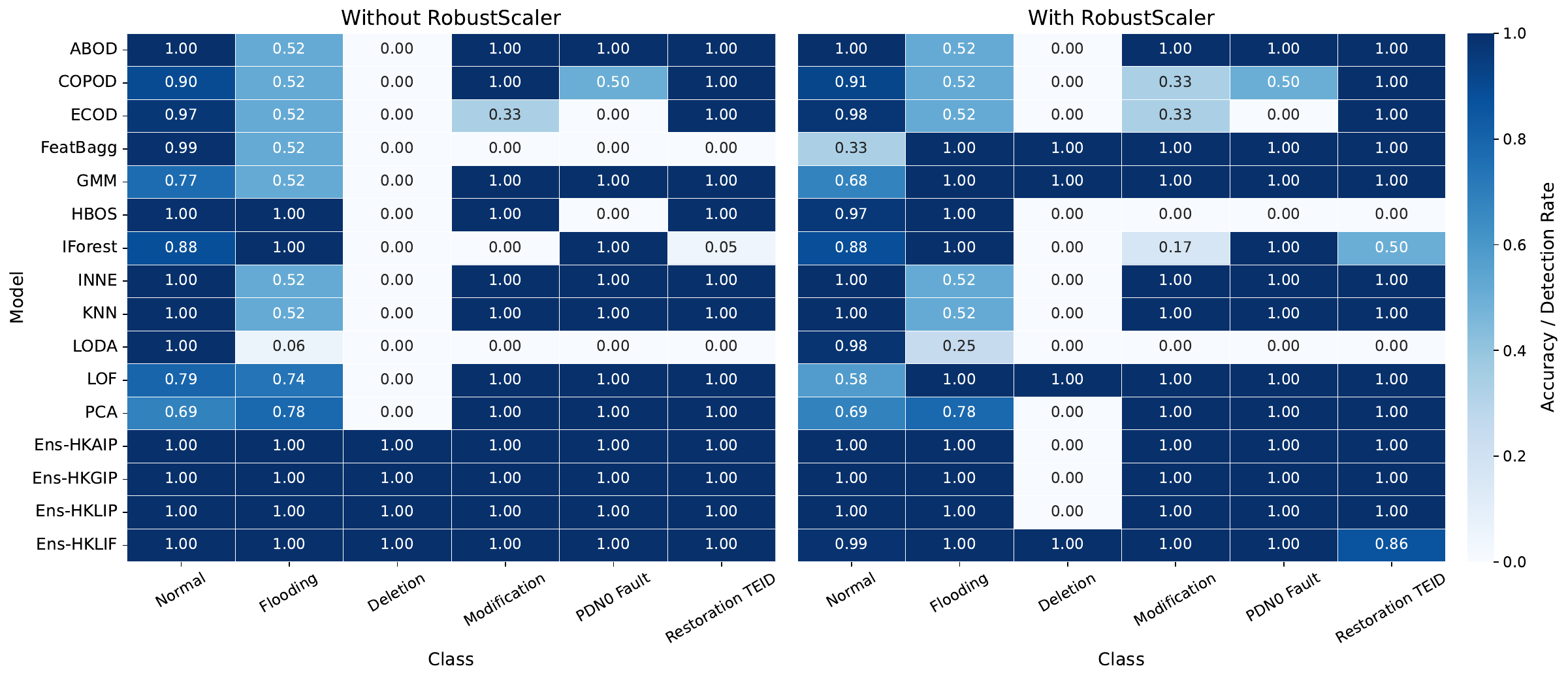}
    \caption{Heat Map of the Detection Rate of the Obtained Models}
    \label{fig:heatmap}
\end{figure}

Evaluations of the detection are summarized in \Cref{tab:overall_performance}, where, \textit{following guideline GE1}, all evaluation metrics are computed consistently across the two settings on the same test set for a fair comparison.
\begin{table}[t]
    \centering
    \begin{tabular}{cc|cccc|cccc}
    & & \multicolumn{4}{c|}{\textbf{Without \texttt{RobustScaler}}} & \multicolumn{4}{c}{\textbf{With \texttt{RobustScaler}}} \\
    \toprule
    & \textbf{Model} & \textbf{AUC} & \textbf{Prec} & \textbf{Recall} & \textbf{F1} & \textbf{AUC} & \textbf{Prec} & \textbf{Recall} & \textbf{F1} \\
    \midrule
    \multirow{3}{*}{\rotatebox{90} {\textbf{Stat.}}}
    & \texttt{HBOS} & \textbf{0.996} & 0.979 & 0.977 & 0.978 & 0.989 & 0.864 & 0.946 & 0.903 \\
    & \texttt{COPOD} & 0.860 & 0.542 & 0.527 & 0.534 & 0.871 & 0.583 & 0.520 & 0.549 \\
    & \texttt{ECOD} & 0.870 & 0.808 & 0.514 & 0.628 & 0.884 & 0.833 & 0.514 & 0.636\\
    \midrule
    \multirow{5}{*}{\rotatebox{90} {\textbf{Density}}}
    & \texttt{kNN} & 0.631 & 0.998 & 0.533 & 0.695 & 0.717 & 0.998 & 0.533 & 0.695 \\
    & \texttt{LOF} & 0.878 & 0.451 & 0.738 & 0.560 & 0.835 & 0.354 & \textbf{1.000} & 0.523 \\
    & \texttt{IForest} & 0.966 & 0.646 & 0.958 & 0.771 & 0.952 & 0.657 & 0.969 & 0.783 \\
    & \texttt{FeatBagg} & 0.772 & 0.953 & 0.489 & 0.647 & 0.844 & 0.256 & \textbf{1.000} & 0.408 \\
    & \texttt{LODA} & 0.370 & \textbf{1.000} & 0.054 & 0.103 & 0.511 & 0.746 & 0.236 & 0.359 \\
    & \texttt{INNE} & 0.640 & 0.998 & 0.532 & 0.694 & 0.850 & \textbf{1.000} & 0.533 & 0.695 \\
    \midrule
    \multirow{3}{*}{\rotatebox{90} {\textbf{Geom.}}}
    & \texttt{PCA} & 0.860 & 0.363 & 0.782 & 0.495 & 0.858 & 0.363 & 0.782 & 0.496 \\
    & \texttt{ABOD} & 0.619 & \textbf{1.000} & 0.533 & 0.695 & 0.749 & 0.997 & 0.533 & 0.694\\
    & \texttt{GMM} & 0.731 & 0.345 & 0.533 & 0.419 & 0.860 & 0.418 & \textbf{1.000} & 0.589 \\
    \midrule
    \midrule
    \multirow{4}{*}{\rotatebox{90}{\textbf{Ensemble}}}
    & \texttt{Ens-HKAIP} & 0.999 & 0.995 & \textbf{1.000} & 0.998 & 0.999 & 0.988 & 0.988 & 0.988\\
    & \texttt{Ens-HKGIP} & 0.999 & 0.995 & \textbf{1.000} & 0.997 & \textbf{1.000} & 0.995 & 0.988 & \textbf{0.992} \\
    & \texttt{Ens-HKLIP} & 0.999 & 0.995 & \textbf{1.000} & 0.998 & \textbf{1.000} & 0.995 & 0.988 & \textbf{0.992} \\
    & \texttt{Ens-HKLIF} & \textbf{1.000} & 0.997 & \textbf{1.000} & \textbf{0.999} & 0.996 & 0.949 & 0.997 & 0.973 \\
    \bottomrule
\end{tabular} 
\caption{Overall performance comparison of anomaly detection models.}
\label{tab:overall_performance}
\end{table}
We show that, among the base detectors, \texttt{HBOS}, \texttt{IForest}, and \texttt{ABOD} achieve the highest F1-scores in the Statistical, Density, and Geometrical categories, respectively. Among the ensemble methods, \texttt{Ens-HKLIP} achieves the best tradeoff between accuracy and F1-score in both its versions, making it the most powerful detector overall in all categories. 
Overall, all ensemble combinations show better overall performance than the individual detectors.
\textit{To provide deeper insight into per-class performance, in accordance with the guideline GE2, we evaluate the detection rates for individual attack categories and normal traffic} using a heatmap, as shown in \Cref{fig:heatmap}. 
This representation highlights the strengths and weaknesses of each approach in recognizing specific types of attacks, as well as their ability to minimize false alarms on benign traffic and identify specific categories of hard-to-detect attacks. Taken together, these evaluation strategies ensure a comprehensive and interpretable assessment of IDS capabilities in realistic, imbalanced 5G attack scenarios.

\subsection{Evaluation of Detectors Under Attack}
\label{sect:results_attacks} 

\textit{After evaluating the performance of the obtained detectors under normal conditions, we assess the robustness under adversarial (feasible and compliant) manipulations, as stated in the guideline GE3.} Specifically, for each model, we generate adversarial samples using both random and optimization-based model-agnostic attack algorithms, as described in \Cref{sec:adv}. In this context, one of the primary evaluation metrics in the adversarial context is the evasion rate, defined as the proportion of adversarially modified attack samples that successfully bypass the detector (i.e., are misclassified as benign). This metric directly quantifies the vulnerability of each IDS to evasion under different attack strategies.
To facilitate comparison, the evasion rates are summarized in \Cref{tab:evasion_rates}. This provides an immediate overview of model robustness and the relative effectiveness of different attack techniques. Results highlight not only the overall susceptibility of the models but also the comparative resilience to simple versus sophisticated adversarial manipulations, supporting a nuanced evaluation of IDS security in a 5G environment. Notably, the results indicate that applying the \texttt{RobustScaler} generally improves robustness against adversarial evasion, particularly under optimization-based model-agnostic attacks, by stabilizing the feature space and reducing the impact of extreme values.

\begin{table}[t]
\centering
\begin{tabular}{cc|rrr|rrr}
& & \multicolumn{3}{c|}{\textbf{Without Scaler}} & \multicolumn{3}{c}{\textbf{With Scaler}} \\
\toprule
& \textbf{Model} 
& \texttt{RS} & \gade & \gaes
& \texttt{RS} & \gade & \gaes \\
\midrule
\multirow{3}{*}{\rotatebox{90}{\textbf{Stat.}}} 
& \texttt{HBOS} & 0.0\% & 98.4\% & 98.4\% & 0.0\% & \textbf{0.0}\% & \textbf{0.0}\% \\
& \texttt{COPOD} & 0.0\% & 0.0\% & 0.0\% & 0.0\% & 0.0\% & 0.0\% \\
& \texttt{ECOD} & 0.0\% & 0.0\% & 0.0\% & 0.0\% & 0.0\% & 0.0\% \\
\midrule
\multirow{5}{*}{\rotatebox{90}{\textbf{Density}}} 
& \texttt{kNN} & 0.0\% & 100.0\% & 100.0\% & 0.0\% & \textbf{0.0}\% & \textbf{0.0}\% \\
& \texttt{LOF} & 0.0\% & 100.0\% & 100.0\% & 0.0\% & 100.0\% & 100.0\% \\
& \texttt{IForest} & 0.0\% & 49.0\% & 0.0\% & 0.0\% & \textbf{0.2}\% & 0.0\%\\
& \texttt{FeatBagg} & 47.1\% & 100.0\% & 100.0\% & \textbf{0.0}\% & \textbf{98.4}\% & \textbf{1.3}\% \\
& \texttt{LODA} & \textbf{5.3}\% & 100.0\% & 100.0\% & 22.6\% & 100.0\% & 100.0\% \\
& \texttt{INNE} & 0.0\% & 100.0\% & 100.0\% & 0.0\% & 100.0\% & 100.0\% \\
\midrule
\multirow{3}{*}{\rotatebox{90}{\textbf{Geom.}}} & 
\texttt{PCA} & 0.0\% & 0.0\% & 0.0\% & 0.0\% & 0.0\% & 0.0\% \\
 & \texttt{ABOD} & 0.0\% & 100.0\% & 100.0\% & 0.0\% & \textbf{0.0}\% & \textbf{0.0}\% \\
 & \texttt{GMM} & 0.0\% & 0.0\% & 0.0\% & 0.0\% & 0.0\% & 0.0\% \\
\midrule
\midrule
\multirow{4}{*}{\rotatebox{90}{\textbf{Ensemble}}} & \texttt{Ens-HKAIP} & 0.0\% & 98.4\% & 98.4\% & 0.0\% & \textbf{50.0}\% & 98.4\% \\
 & \texttt{Ens-HKGIP} & 0.0\% & 98.4\% & 98.4\% & 0.0\% & \textbf{50.0}\% & \textbf{48.7}\% \\
 & \texttt{Ens-HKLIP} & 0.0\% & 98.4\% & 98.4\% & 0.0\% & \textbf{50.0}\% & \textbf{48.7}\% \\
 & \texttt{Ens-HKLF} & 0.0\% & 100.0\% & 100.0\% & 0.0\% & 100.0\% & 100.0\% \\
\bottomrule
\end{tabular}
\caption{Evasion rates under random and adaptive model-agnostic attack strategies.}
\label{tab:evasion_rates}
\end{table}

\section{Related Work}
\label{sec:related}
Our work lies at the intersection between anomaly detection and adversarial machine learning in the 5GC network, with an explicit focus on realistic and practical approaches to the underlying attack problem. We thereby discuss the related work for each of these topics.

\myparagraph{ML-based intrusion and anomaly detection in the 5GC.}
A growing body of work studies ML methods to monitor and secure the 5GC, motivated by the complexity and volume of control- and user plane-traffic. 
Early work has explored the application of ML for anomaly detection in emerging 5G networks, primarily from a flow-level perspective, demonstrating the feasibility of ML-based detection for 5G traffic and highlighting its potential to identify malicious traffic in yet another application scenario~\cite{lam2020machine}.
Recent approaches propose ML-powered Intrusion Detection System (IDS) pipelines tailored to 5GC interfaces and protocols, often emphasizing PFCP-related threats or focusing on explainability-based techniques to support operator decision making. 
For instance, 5GCIDS introduces an AI-based IDS specifically targeting PFCP and discusses the use of explainability-based techniques as an aid for security monitoring~\cite{radoglou20235gcids}.
Related work has also investigated the design of learning-based intrusion detection pipelines for 5GC with a particular focus on efficiency and scalability~\cite{kim2022effective}.
In contrast to prior work, which evaluates ML-based 5GC intrusion detection under static and non-adversarial assumptions, we focus on realistic deployment scenarios and propose evaluation guidelines that explicitly consider adaptive attackers manipulating protocol-related traffic to evade detection. 

\myparagraph{Adversarial machine learning in the 5GC.}
Closer to our work, prior research has highlighted that 5G management and security tasks can be exposed to adversarial manipulation when ML components are deployed in operational pipelines. Specifically, in~\cite{apruzzese_22_wild}, the authors study this specific scenario, emphasizing attacker constraints and threat models with adversaries and ML-based defenses.
However, the focus of this line of work is on adversarial interference with ML systems that support radio access and network management tasks (e.g., slicing, CQI prediction, modulation recognition), rather than on evading ML-based anomaly detection systems operating on 5GC protocol traffic. In contrast, our work specifically addresses anomaly detection in the 5G Core control plane and complements existing research by providing security-aware evaluation guidelines that explicitly account for adaptive attackers capable of manipulating protocol-compliant traffic to evade detection in realistic deployment scenarios. 

\myparagraph{Assumptions, guidelines, and realistic attacks in ML-based detection.}
Recent systematization and survey works have highlighted that many ML-based network traffic analysis systems rely on fragile design choices and implicit assumptions—such as environment-dependent identifiers, legacy datasets, or shortcut features—that can lead to overfitting and limited generalization in real deployments~\cite{wickramasinghe2025sok}.
Similar concerns regarding evaluation realism and deployment-oriented assumptions have been raised in recent methodological analyses of learning-based network security systems, further motivating the need for principled and security-aware evaluation practices~\cite{zhao2025sweet}.
Our work also builds on these observations, extending and instantiating their methodological principles in the context of 5GC. In particular, we translate high-level concerns about feature dependence and evaluation realism into concrete, protocol-aware guidelines for anomaly detection in 5GC environments, and further develop them by explicitly considering adversaries that manipulate protocol-compliant traffic to evade learning-based detectors.
From a different standpoint, a line of work argues for realistic problem-space perturbations and demonstrates that such attacks can effectively degrade NIDS performance under practical constraints~\cite{catillo2024towards}. 
In contrast, our work does not involve traffic-space manipulation but adopts a complementary and deployment-oriented threat model for the 5GC. Hence, rather than arbitrary feature perturbations, we study adversarial evasion at the feature level under strict protocol and attack constraints, explicitly modeling the subset of related features that an attacker can realistically control without breaking protocol compliance or system functionality.

\section{Conclusion}
\label{sec:conclusion}
In this work, we addressed the gap between current evaluation methodologies for anomaly detection in the 5GC network and the operational reality of deployed networks. We introduced \textbf{SAGE-5GC}, a comprehensive set of guidelines for security-aware detection that prioritizes model independence and, crucially, robustness against adaptive attacks.

Through a comprehensive experimental campaign using realistic datasets for 5GC traffic, we demonstrated that high detection performance does not align with its security in the wild. While ensemble-based detectors achieved near-perfect accuracy against static attacks, our evaluation revealed their fragility under adversarial conditions. By employing constrained, model-agnostic optimization strategies (specifically random search and genetic algorithms) we showed that an attacker can generate protocol-compliant perturbations that successfully evade detection without compromising the efficacy of the attack.
These findings underscore that standard metrics such as AUC and F1-score are insufficient for security-critical applications when used in isolation. We conclude that future evaluations of 5G anomaly detection systems must explicitly incorporate adversarial robustness assessments.
Adopting the SAGE-5GC guidelines provides a necessary foundation for developing next-generation detectors that are not only accurate but also resilient to the evolving threat landscape of 5GC networks.\medskip

\myparagraph{Limitations and Future Work.}
Despite the significance of these findings, our study has limitations. Our adversarial evaluation relies on a manually defined set of feasible features $J$ based on domain knowledge; however, in complex, multi-vendor environments, identifying these constraints may require automated inference. Furthermore, our analysis focuses exclusively on the PFCP protocol, leaving vulnerabilities related to other protocols unexplored.
Future work will aim to bridge these gaps. Therefore, we plan to optimize the set of feasible features $J$ leveraging discrete optimization techniques. 
Finally, while our adversarial are feasible and compliant, they are not reconstructed fully to replayable and fully-valid network traffic. 
Generating adversarial samples at the packet or flow level to produce executable traffic traces that can be validated in realistic testbeds.

\section*{Acknowledgments} 
This work was partially supported by the EU-funded project Sec4AI4Sec (grant no. 101120393); and by the project SERICS (PE00000014) under the MUR NRRP funded by EU-NextGenEU. This work was carried out while C. Scano was enrolled in the Italian National Doctorate on AI run by the Sapienza University of Rome in collaboration with the University of Cagliari.

\bibliographystyle{unsrt}
\bibliography{bibliography}
\clearpage
\appendix
\begin{center}
  \LARGE
  {Supplementary Material of \\``\ourtitle''}
\end{center}
\section{Further Experimental Details}
This section provides additional details on the experimental settings.
\subsection{Datasets}\label{app:datasets}
The various cyberattacks implemented are:
\myparagraph{PFCP Restoration-TEID.} The PFCP Restoration-TEID attack (CVE-2025-29646) disrupts the PFCP session recovery process by injecting forged restoration messages containing manipulated TEID values. The attack specifically targets \texttt{pfcp.f\_teid.teid}, setting it to values exceeding the TEID pool size (TEID $>$ 1024 × 4 × 16 = 65536), alongside manipulated \texttt{pfcp.f\_teid.ipv4\_addr}, \texttt{pfcp.pdr\_id}, and \texttt{pfcp.node\_id\_ipv4} fields. Additional fields modified include \texttt{pfcp.ie\_type}, \texttt{pfcp.ie\_len}, \texttt{pfcp.msg\_type}, \texttt{pfcp.f\_\-seid.ipv4}, \texttt{pfcp.flags}, the session flag \texttt{pfcp.s}, \texttt{pfcp.seid}, and the sequence number \texttt{pfcp.se\-qno}. Since TEIDs bind PFCP control-plane state to user-plane tunnel identifiers, corrupting them causes the UPF to reconstruct incorrect forwarding state, leading to traffic misrouting, session drops, or UPF crash.  The attack exploits the lack of authentication and integrity checks in PFCP recovery procedures, allowing an attacker capable of spoofing PFCP packets to desynchronize the UPF's control-plane and user-plane state.\medskip

\myparagraph{PFCP Flood.} The PFCP Flood attack overwhelms PFCP control-plane entities with a high volume of unsolicited PFCP messages.  The attack manipulates \texttt{pfcp.msg\_type} (primarily Session Establishment Requests with value 50 and Heartbeat Requests), randomizes \texttt{pfcp.seqno} and \texttt{pfcp.ue\_ip\_addr\_ipv4}, while also modifying \texttt{pfcp.seid} and \texttt{pfcp.s} flag unnecessarily. Because PFCP operates over UDP (port 8805) without built-in rate limiting, the receiving UPF or SMF is forced to process excessive traffic, exhausting control-plane resources and delaying or blocking legitimate signaling. This results in reduced session management performance and potential denial of service.

\myparagraph{PFCP Deletion.} In the PFCP Deletion attack, an attacker sends forged Session Deletion Requests by manipulating \texttt{pfcp.msg\_type} (set to 54), \texttt{pfcp.s} flag, and targeting active \texttt{pfcp.seid} values to prematurely remove session state at the UPF. Additional fields modified include \texttt{pfcp.seqno}, \texttt{pfcp.length}, and \texttt{pfcp.flags}. This forces the immediate deletion of PDRs, FARs, and associated tunnel information, interrupting user-plane flows and causing abrupt service disruption. The attack is feasible because PFCP does not authenticate control messages, making it possible for a spoofing attacker to trigger unauthorized session teardown.\medskip

\myparagraph{PFCP Modification.} The PFCP Modification attack injects counterfeit Session Modification messages (\texttt{pfcp.msg\_type = 52}) targeting active sessions identified by their \texttt{pfcp.seid} values. The attack operates by altering Forwarding Action Rules (FARs) within the UPF:  specifically, it sets \texttt{pfcp.apply\_action.forw} to 0 (disabling forwarding) while setting \texttt{pfcp.apply\_action.bu\-ff} and \texttt{pfcp.apply\_action.nocp} to manipulate buffering and notification behavior, effectively converting legitimate forwarding rules into DROP actions. Simultaneously, the attack corrupts forwarding parameters by modifying \texttt{pfcp.outer\_hdr\_creation.ipv4} (destination IP for encapsulated traffic), \texttt{pfcp.outer\_hdr\_creation.teid} (often set to invalid values to cause tunneling failures), and \texttt{pfcp.dst\_interface}, redirecting packets to incorrect network interfaces. These changes are complemented by modifications to protocol-level fields such as \texttt{pfcp.s}, \texttt{pfcp.seqno}, \texttt{pfcp.length}, and \texttt{pfcp.flags}. By systematically corrupting these forwarding and encapsulation parameters, the attacker can cause user traffic to be dropped, misrouted to the wrong network segments, or encapsulated with invalid tunnel headers that result in packet discard. The attack exploits PFCP's lack of message authentication and integrity protection, allowing unauthorized modification of critical control-plane state that directly governs user-plane traffic processing at the UPF.\medskip

\myparagraph{UPF PDN-0 Fault.} The UPF PDN-0 Fault attack exploits weaknesses in the UPF's handling of PDN Type 0 session context (\texttt{pfcp.pdn\_type=0}). The attack manipulates key session establishment fields including \texttt{pfcp.node\_id\_ipv4}, \texttt{pfcp.f\_seid.ipv4}, \texttt{pfcp.ue\_ip\_addr\_ipv4} and \texttt{pfcp.pdr\_id}, creating an inconsistent state.  It also sets F-TEID flags to invalid combinations:  \texttt{pfcp.f\_teid\_flags.ch}, \texttt{pfcp.f\_teid\_flags.ch\_id}, and \texttt{pfcp.f\_teid\_flags.v6}. Additional modified fields include \texttt{pfcp.msg\_type}, \texttt{pfcp.flags}, \texttt{pfcp.s}, and \texttt{pfcp.seqno}. This attack induces state inconsistencies that can cause session establishment failures, unexpected teardown, or misrouting of user traffic, highlighting insufficient validation of PFCP parameters in PDN context management.\medskip
\end{document}